\documentclass[10pt,twocolumn,letterpaper]{article}

\usepackage{spconf}
\usepackage{times}
\usepackage{epsfig}
\usepackage{graphicx}
\usepackage{amsmath}
\usepackage{amssymb}
\usepackage{siunitx}
\usepackage{xcolor}
\usepackage{multirow}
\usepackage{subcaption}
\usepackage{threeparttable}
\usepackage{algorithm}
\usepackage{algorithmic}
\usepackage{makecell}
\graphicspath{ {images/} }
\usepackage{tabularx}

\usepackage[pagebackref=true,breaklinks=true,letterpaper=true,colorlinks,bookmarks=false]{hyperref}

\begin{document}
\ninept  
\newcommand\todo[1]{\textcolor{red}{#1}}
\title{Shift R-CNN: Deep Monocular 3D Object Detection with Closed-Form Geometric Constraints}

%

\makeatletter
\renewcommand*{\@fnsymbol}[1]{\ensuremath{\ifcase#1\or \dagger\or \ddagger\or
    \mathsection\or \mathparagraph\or \|\or **\or \dagger\dagger
    \or \ddagger\ddagger \else\@ctrerr\fi}}
\makeatother

\makeatletter
\newcommand{\printfnsymbol}[1]{%
  \textsuperscript{\@fnsymbol{#1}}%
}
\makeatother

\makeatletter
\def\ps@IEEEtitlepagestyle{%
  \def\@oddfoot{\mycopyrightnotice}%
  \def\@evenfoot{}%
}
\def\mycopyrightnotice{%
  {\hfill \footnotesize \copyright 2019 IEEE. Personal use of this material is permitted. Permission from IEEE must be obtained for all other uses, in any current or future media, including reprinting/republishing this material for advertising or promotional purposes, creating new collective works, for resale or redistribution to servers or lists, or reuse of any copyrighted component of this work in other works.\hfill
  \\
  
  Accepted to be published in 2019 IEEE International Conference on Image Processing, Sep 22-25, 2019, Taipei.
  }
}
\makeatother

\twocolumn[
\begin{@twocolumnfalse}
    \mycopyrightnotice
\end{@twocolumnfalse}
]

\name{Andretti Naiden\sthanks{A. Naiden and V. Paunescu contributed equally to this work.}$^{1}$, Vlad Paunescu\printfnsymbol{1}$^{1}$, Gyeongmo Kim$^{2}$, ByeongMoon Jeon$^{2}$, Marius Leordeanu$^{1,3}$}


\address{$^{1}$Arnia Software, Bucharest, Romania\\
$^{2}$Advanced Camera Laboratory, LG Electronics, Seoul, Korea\\
$^{3}$Politehnica University of Bucharest, Bucharest, Romania
}

%
%
%

\maketitle

\begin{abstract}

We propose Shift R-CNN, a hybrid model for monocular 3D object detection, which combines deep learning with the power of geometry. We adapt a Faster R-CNN network for regressing initial 2D and 3D object properties and combine it with a least squares solution for the inverse 2D to 3D geometric mapping problem, using the camera projection matrix.  The closed-form solution of the mathematical system, along with the initial output of the adapted Faster R-CNN are then passed through a final ShiftNet network that refines the result using our newly proposed \textit{Volume Displacement Loss}. Our novel, geometrically constrained deep learning approach to monocular 3D object detection obtains top results on KITTI 3D Object Detection Benchmark \cite{kitti-3d}, being the best among all monocular methods that do not use any pre-trained network for depth estimation.
 
\end{abstract}


\section{Introduction}
Autonomous driving relies on visual perception of the surrounding environment, in which the ability to perform accurate obstacle detection is a key factor. With many different types of sensors available, state of the art autonomous driving systems are based on sensor fusion \cite{avod}. When compared to LiDAR systems, which have a higher cost of production, RGB cameras are ubiquitous, much less expensive, and have a significantly higher pixel resolution. Humans drive using only visual cues, therefore high performance autonomous driving could be achieved, in principle, using vision alone. 

While stereo cameras are able to recover the depth of the scene, monocular 3D perception still remains a very difficult challenge. In this paper we focus our efforts on monocular 3D object detection and propose an efficient method that combines the initial Faster R-CNN predictions with a closed-form, 2D to 3D geometric mapping solution that produces a rough estimation of an object's 3D translation. This translation is then refined with our proposed ShiftNet network.

\begin{figure*}[t]
\begin{center}
\includegraphics[width=1.0\linewidth]{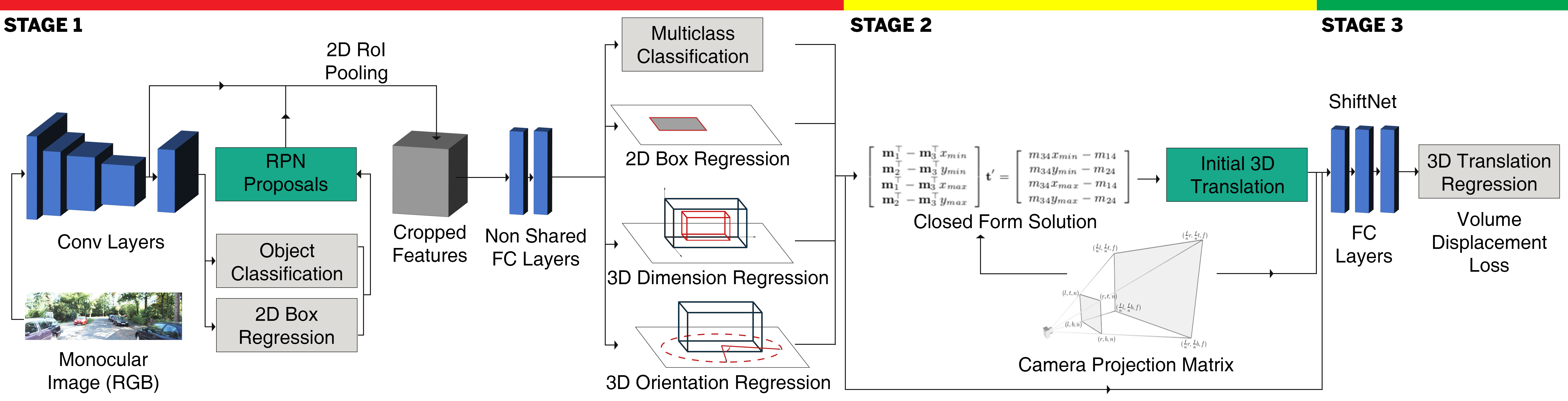}
\end{center}
   \caption{Overview of our Shift R-CNN hybrid model. Stage 1: Faster R-CNN with added 3D angle and dimension regression. Stage 2: Closed-form solution to 3D translation using camera projection geometric constraints. Stage 3: ShiftNet refinement and final 3D object box reconstruction.
   }
\label{fig:our_3stage_system}
\end{figure*}

In summary, the main contributions of our paper are:
\begin{enumerate}
\item A 3D object detection architecture with three stages, that accurately estimates the 2D location, 3D object properties (height, width, length), its orientation and 3D translation in camera coordinates, from a single RGB image (Fig. \ref{fig:our_3stage_system}). In the first stage, an adapted Faster R-CNN estimates the 2D box, dimension and orientation. In the second stage, a mathematical system of equations is solved using least squares, based on the previous estimates and the known camera projection matrix, to efficiently compute the 3D object translation by enforcing the 3D to perfectly fit the 2D. In the third stage, a fully connected network, namely ShiftNet, learns the error dependency between the first 2 stages and corrects the final translation.
\item A novel \textit{Volume Displacement Loss (VDL)} used to train a network for the task of maximizing the 3D IoU by means of regression.
\item State of the art results on the KITTI 3D Object Detection Benchmark \cite{kitti-3d}, compared to all other published monocular methods that do not make use of an already trained depth net.
\end{enumerate}

\noindent \textbf{Related Work.} There is a vast literature on 2D object detection, with accurate approaches on challenging datasets. They vary from general ones, such as COCO \cite{deataset:coco}, to those dedicated to autonomous driving: KITTI \cite{kitti-3d}, Baidu Apollo \cite{dataset:apollo}, and Berkley DeepDrive \cite{dataset:bdd100k}. 2D object detection architectures can be divided into ROI-based, two-stage models and proposal free, single-shot approaches. Single-shot detectors include YOLO \cite{yolo}, SSD \cite{ssd} and RetinaNet \cite{retinanet}. Two-stage detectors include Faster R-CNN \cite{fasterrcnn}, Mask R-CNN \cite{maskrcnn}, PANet \cite{panet}. 


The most powerful 3D object detection models rely on LiDAR-generated point clouds, which simplify the 3D estimation task. 3D-FCN \cite{3dfcn} and VoxelNet \cite{voxelnet} apply 3D convolutions to point-cloud data to generate 3D bounding boxes. AVOD \cite{avod} uses sensor fusion and jointly predicts 3D properties from LiDAR and RGB, being robust to noisy LiDAR data. PointRCNN \cite{pointrcnn} performs a point cloud foreground/background segmentation, then pools semantic point clouds for accurate canonical 3D object detection. Other approaches use stereo images for a good trade-off between quality and price. Top methods in this category use depth generated from stereo as well as RGB input. Methods include \cite{3dop-stereo,3dop2,DeepStereoOP}, which use stereo information to create 3D object proposals that are scored with CNNs. 

Monocular 3D object detection is the most difficult task since it requires tackling the inverse geometrical problem of mapping 2D space into 3D space, in the absence of any true 3D information. Thus, top approaches rely on extra training data in order to make informed 3D estimations. Wang \textit{et al.} \cite{pseudo-lidar} use monocular depth perception networks such as DORN \cite{dorn} to generate pseudo-point clouds and then apply a state of the art LiDAR-based model \cite{avod}. ROI-10D \cite{roi_10d} uses monocular depth estimation and lifts 2D detection into a 6D pose problem. Mono3D \cite{mono3d} enforces 3D candidates to lay on the ground plane, orthogonal to image plane. It also uses semantic segmentation, contextual information, object size, shape and location priors. Deep3DBox \cite{3d-deepbox} uses geometric constraints by tightly fitting the 3D bounding box into the 2D box. Xu \textit{et al.} \cite{mlf-mono} use multi-level fusion of Faster R-CNN and depth features for 3D object detection. Advances in stereo and monocular depth estimation could provide accurate 3D information, which could greatly improve non-LiDAR based 3D object detection systems.

\section{Our 3D Object Detection Architecture}
In the absence of any true depth information, estimating the 3D object translation from a single image is a harder problem than estimating its 2D bounding box, dimension and orientation. Our Shift R-CNN model (Fig. \ref{fig:our_3stage_system}), combining deep learning with geometric constraints is mainly based on this observation.

\subsection{Stage 1: 2D Object Detection and 3D Intrinsics Estimation}
\label{sec:stage1}
The first stage augments the Faster R-CNN model with additional learning objectives. First, we extract and score 2D region proposals by means of 2D \textit{anchors} \cite{fasterrcnn}, using objectness classification and 2D box regression. We then employ 2D ROI pooling for feature cropping. Based on the top scoring 2D proposals, we refine the cropped features using a convolutional encoder, then split them up into 4 separate heads. For the 2D part we use standard multi-class classification and 2D box refinement. The additional 2 heads that we introduce handle local orientation and 3D dimension regression. 

\noindent\textbf{Angle Regression Head.} Due to the periodic nature of angles, it is harder to regress them explicitly. Thus, we regress instead the estimates $\widehat{\sin_{\alpha_L}}$ and $\widehat{\cos_{\alpha_L}}$ of the local orientation angle $\alpha_L$ w.r.t the y-axis, by also adding a constraint loss to enforce  
$\widehat{\sin_{\alpha_L}}^2 + \widehat{\cos_{\alpha_L}}^2 = 1$. We predict for each object $C$ different angles, one for each class, and compute the loss only on the winning one w.r.t the ground-truth local angle $\alpha_{L_{gt}}$. 
\begin{equation}
    L_{\alpha_L} = ||\sin{\alpha_{L_{gt}}} - \widehat{\sin_{\alpha_L}}||_2  + ||\cos{\alpha_{L_{gt}}} - \widehat{\cos_{\alpha_L}}||_2 + L_{cnt}, 
\end{equation}
\begin{equation}
    L_{cnt}= ||1 - (\widehat{\sin_{\alpha_L}}^2+\widehat{\cos_{\alpha_L}}^2)||_{2}.
\end{equation}
At inference time, $\alpha_L = atan2(\widehat{\sin_{\alpha_L}}, \widehat{\cos_{\alpha_L}})$.\\
\noindent\textbf{3D Dimension Regression Head.} The 3D object height, width and length (in meters) are also regressed in $C$ triplets, one per class. We do not regress the absolute dimensions, due to their low variance per class. We adopt a procedure similar to 2D bounding box regression \cite{fasterrcnn}, by taking into account predefined mean dimensions. Let the mean dimension for a particular class $c$ be $\overline{\mathbf{d}_c} =\left[ {\begin{array}{ccc}
   \overline{h_c}, \overline{w_c}, \overline{l_c}\\
  \end{array} } \right]$, we denote $\mathbf{d} = [h, w, l]$ as the object's real predicted dimension and regress the logarithmic scale offsets $\Delta \mathbf{d}_c = [\Delta h, \Delta w, \Delta l]$ w.r.t $\overline{\mathbf{d}_c}$. The relation between $\Delta \mathbf{d}_c$ and $\mathbf{d}$ is seen in Eq. \ref{eq:dims_reg}. The loss is computed only for the winner class.
\begin{equation}
    \Delta \mathbf{d}_c = \left[{\begin{array}{ccc} \ln{\frac{h}{\overline{h_c}}} , \ln{\frac{w}{\overline{w_c}}} , \ln{\frac{l}{\overline{l_c}}} \end{array}} \right],
\end{equation}

\begin{equation}
\label{eq:dims_reg}
    \mathbf{d} = [e^{\Delta h} \times \overline{h_c},
    e^{\Delta w} \times \overline{w_c},
    e^{\Delta l} \times \overline{l_c}].
\end{equation}

\noindent\textbf{Multi-Task Loss.} We present our total Stage 1 loss in Eq. \ref{eq:total_loss}. The first 2 terms are the standard Faster R-CNN multiclass loss and 2D bounding box losses. For the 3D dimension regression we use a standard $L_2$ loss, $L_{dim}$, over the logarithmic scale offsets $\Delta \mathbf{d}_c$. 
\begin{equation}
\label{eq:total_loss}
    L_{total}= w_{cls} L_{cls} + w_{2D} L_{2D} +  w_{\alpha_L} L_{\alpha_{L}} + w_{dim} L_{dim}.
\end{equation}
\subsection{Stage 2: Estimating the 3D Object Translation}
 Using the 2D bounding box $\mathbf{b}_{2D} = [x_{min}, y_{min}, x_{max}, y_{max}]$, the 3D dimension estimates $\mathbf{d} = [h, w, l]$ and the local orientation angle about the y-axis $\alpha_L$, the 3D bounding box reconstruction follows. \\
 
 \begin{figure*}[t]
\begin{center}
\includegraphics[width=1.0\linewidth]{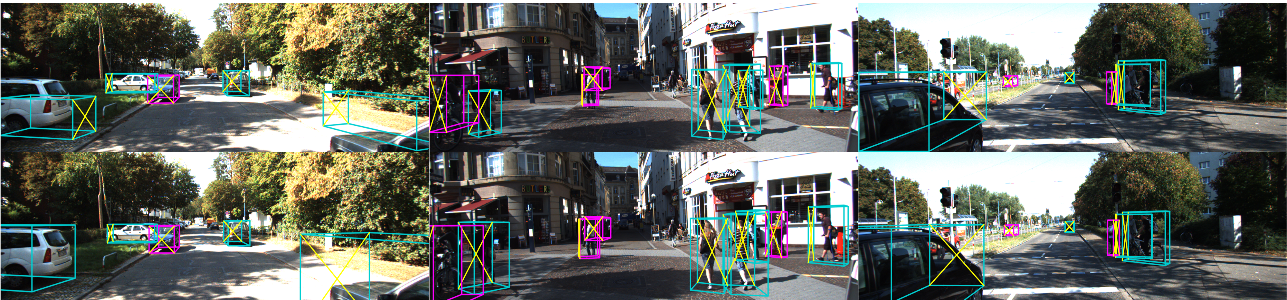}
\end{center}
   \caption{Stage 2 (top) and Stage 3 (bottom) results comparison. Note that Stage 3 improves the 3D estimation due to its noise robustness. Turquoise boxes denote objects with the same orientation and magenta color the opposite orientation. Best viewed in color.}
\label{fig:ablation}
\end{figure*}
 
\noindent\textbf{From local to global orientation.}
Having already predicted the local orientation angle $\alpha_L$, the global orientation angle $\alpha_G$ is then determined by considering the ray from the camera origin to the center of the object $\theta_{ray}$: \begin{equation}
\label{eq:theta_ray}
 \alpha_G = \alpha_L - \theta_{ray}.
\end{equation} 

\noindent\textbf{Inverse geometry problem.} In order to accurately reconstruct the 3D bounding box, we need to further estimate the real 3D translation vector $\mathbf{t} = [t_x, t_y, t_z]^\top$ of the object center, in 3D camera coordinates. We formulate this estimation in a closed-form, as a least squares solution given by fitting the \textit{geometric constraints} imposed by the camera projection matrix $\mathbf{P}$. Thus, an object described by $\mathbf{b}_{2D}$, $\mathbf{d}$ and $\alpha_G$, will have a depth-constrained translation $\mathbf{t}' = [t'_x, t'_y, t'_z]^\top$ in camera coordinates.

To enforce the 3D bounding box projection to fit tightly into the predicted 2D bounding box, we constrain 2 of the 4 vertical 3D edges to lay on a 2D vertical side and the upper and lower 3D corners to lay on a horizontal 2D side. Assuming that objects lay on the ground plane and that by fixing one vertical 3D edge the second vertical one must be diagonally opposed, we have $4 \times 1 \times 4 \times 4 = 64$ configurations - from which we choose the best fit. We reconstruct the 3D bounding box $\mathbf{x}_{3D_o}$ at the camera center, by using the 3D dimensions $\mathbf{d}$, then obtain the final 3D box, in camera coordinates, by applying the rotation about the y-axis $\mathbf{R}_y(\alpha_G)$ and translation $\mathbf{t}'$:
\begin{equation}
\label{eq:translation}
 \mathbf{x}_{3D} =  \mathbf{R}_y(\alpha_G) \mathbf{x}_{3D_o} + \mathbf{t}'.
\end{equation} 

The relation between a 3D point in the world $\mathbf{x}_{3D} = [x, y, z]^\top$ and its 2D projection in the image $\mathbf{x}_{2D} = [x, y]^\top$, using the projection matrix $\mathbf{P}$, is given in homogeneous coordinates:
\begin{equation}
\label{eq:projection}
   \lambda \left[\begin{array}{c}
        \mathbf{x}_{2D}  \\
        1 \end{array}
        \right] = \mathbf{P} \left[\begin{array}{c}
        \mathbf{x}_{3D}  \\
        1 \end{array}
        \right].
\end{equation}

From Eq. \ref{eq:translation} and \ref{eq:projection}, similar to \cite{3d-deepbox},
we obtain the following system where translation $\mathbf{t'}$ is the unknown vector:
\begin{equation}
\label{eq:system}
    \underbrace{ \mathbf{P} \left[\begin{array}{cc}
        I & \mathbf{R}_y(\alpha_G) X_{3D_o} \\
        0 & 1 \end{array} \right] }_\textit{$\mathbf{M}$}
        \left[\begin{array}{c}
        \mathbf{t}'  \\
        1 \end{array} \right] =
        \lambda \left[\begin{array}{c}
        x_{2D_{side}}  \\
        y_{2D_{side}}  \\
        1 \end{array}\right].
\end{equation}

By substituting each element of $\mathbf{b}_{2D}$, corresponding to a 2D side, and also $\lambda$ into System \ref{eq:system}, we propose a least squares solution for the translation, different from \cite{3d-deepbox}. Here $\mathbf{m_i}^\top = [m_{i1},m_{i2},m_{i3}]$ and $m_{ij}$ is the $(i,j)$ element of matrix $\mathbf{M}$.
\begin{equation}
\label{eq:system_explicit}
    \left[\begin{array}{c}
        \mathbf{m}_1^\top - \mathbf{m}_3^\top x_{min} \\
        \mathbf{m}_2^\top - \mathbf{m}_3^\top y_{min} \\
        \mathbf{m}_1^\top - \mathbf{m}_3^\top x_{max} \\
        \mathbf{m}_2^\top - \mathbf{m}_3^\top y_{max} 
        \end{array}
        \right] \mathbf{t}'  
         =  \left[\begin{array}{c}
        m_{34}x_{min} - m_{14} \\
        m_{34}y_{min} - m_{24} \\
        m_{34}x_{max} - m_{14} \\
        m_{34}y_{max} - m_{24} 
        \end{array}
        \right].
\end{equation}

The over-constrained System \ref{eq:system_explicit} can be rewritten as $\mathbf{A} \mathbf{t}' = \mathbf{b}, \mathbf{b} \neq \mathbf{0}$, with a general closed-form solution for the 3D object translation $\mathbf{t}' = (\mathbf{A}^\top \mathbf{A})^{-1}\mathbf{A}^\top \mathbf{b}$. For choosing the best fitting solution, we evaluate the 2D IoU score of the initial 2D and re-projected 3D box.  

\subsection{Stage 3: Refining the 3D Object Translation} 
\textbf{Shift Network.} The $\mathbf{t}' = [t'_x, t'_y, t'_z]^\top$ translation estimate from Eq. \ref{eq:system_explicit} alone is a good approximation of the object 3D location. However, it assumes a perfect mathematical model and, implicitly, perfect Faster R-CNN predictions. Our tests indicate that the estimate is not accurate when predictions are noisy, going from 80\% accuracy with a 3D IoU of at least 0.7, to a roughly 14\% if noise is present. Even a small error in the predicted values needed to solve Eq. \ref{eq:system_explicit}, produce huge displacements of $\mathbf{t}'$ and thus low 3D IoU score.

To address this limitation, we propose an effective approach for
refining the estimate $\mathbf{t}'$. We introduce \textit{ShiftNet} (SHN) in Stage 3. We model ShiftNet as a hybrid that uses the $\mathbf{t}'$ estimate from Stage 2 (Eq. \ref{eq:system}), with the initial Stage 1 predictions and outputs an improved final translation $\mathbf{t}'' = [t''_x, t''_y, t''_z]$. It learns to combine, in complementary ways, the best of the exact geometric universe with the ability of a neural net to learn in the presence of noise. The actual input fed to the network is given by $\mathbf{t}'$, $\mathbf{b}_{2D}$, $\mathbf{d}$, $(sin(\alpha_L), cos(\alpha_L))$, $(sin(\alpha_G), cos(\alpha_G))$ and $\mathbf{P}$. SHN is then trained to regress $\mathbf{t}''$. The architecture has 2 fully-connected layers with 1024 neurons, and a final layer that regresses 3 values for $\mathbf{t}'' = [t''_x, t''_y, t''_z]$.\\

\begin{table*}[t]
\centering
\resizebox{\textwidth}{!}{\begin{tabular}{|c|c|c|c|c|c|c|c|c|}
\hline
 \multicolumn{3}{|c}{} & \multicolumn{3}{|c}{$AP_{3D}$ (\%)} & \multicolumn{3}{|c|}{$AP_{BEV}$ (\%)} \\ \hline 
Method & Setup & Class & Easy & Moderate & Hard & Easy & Moderate & Hard \\ \hline 

Mono3D \cite{mono3d}& Mono & \multirow{7}{*}{\makecell{Car \\ (IoU$\geq$0.7)}} & 
\,\,\,2.53 / \,\,\,\,\,\,-\,\,\,\,\,\, &
\,\,\,2.31 / \,\,\,\,\,\,-\,\,\,\,\,\, &
\,\,\,2.31 / \,\,\,\,\,\,-\,\,\,\,\,\, &
\,\,\,5.22 / \,\,\,\,\,\,-\,\,\,\,\,\, & 
\,\,\,5.19 / \,\,\,\,\,\,-\,\,\,\,\,\, & 
\,\,\,4.13 / \,\,\,\,\,\,-\,\,\,\,\,\,
\\

DeepBox3D \cite{3d-deepbox}  & Mono & &
\,\,\,\,\,\,-\,\,\,\,\,\, / \,\,\,5.85  &
\,\,\,\,\,\,-\,\,\,\,\,\, / \,\,\,4.10 &
\,\,\,\,\,\,-\,\,\,\,\,\, / \,\,\,3.84 &
\,\,\,\,\,\,-\,\,\,\,\,\, / \,\,\,9.99 &
\,\,\,\,\,\,-\,\,\,\,\,\, / \,\,\,7.71 &
\,\,\,\,\,\,-\,\,\,\,\,\, / \,\,\,5.30  
\\

OFT-Net \cite{oft-net}  & Mono & & 
\,\,\,4.07 / \,\,\,2.50 &
\,\,\,3.27 / \,\,\,3.28 &
\,\,\,3.29 / \,\,\,2.27 &
11.06 / \,\,\,9.50 &
\,\,\,8.79 / \,\,\,7.99 &
\,\,\,8.91 / \,\,\,\textbf{7.51} 
\\ 

\color{blue}MLF-Mono \cite{mlf-mono}*  & \color{blue}Mono+PD & & 
\color{blue}10.53 / \,\,\,7.85 &
\color{blue}\,\,\,5.69 / \,\,\,5.39 &
\color{blue}\,\,\,5.39 / \,\,\,4.73 & 
\color{blue}22.03 / 19.20 &
\color{blue}13.63 / 12.17 &
\color{blue}11.60 / 10.89
\\
\color{blue}ROI-10D \cite{roi_10d}* & \color{blue}Mono+PD & & 
\color{blue}10.25 / 12.30 &
\color{blue}\,\,\,6.39 / 10.30 &
\color{blue}\,\,\,6.18 / \,\,\,9.39 &
\color{blue}14.76 / 16.77 &
\color{blue}\,\,\,9.55 / 12.40 &
\color{blue}\,\,\,7.57 / 11.39 
\\
Linear System (Ours)  & Mono & & 
\,\,\,7.24 / \,\,\,6.80 &
\,\,\,5.98 / \,\,\,4.14 &
\,\,\,5.54 / \,\,\,3.50 &
14.74 / 11.75  &
12.48 / \,\,\,8.34 &
11.22 / \,\,\,6.80
\\

ShiftNet (Ours)  & Mono & & 
\,\,\,\textbf{\color{blue}13.84} / \,\,\,\,\textbf{8.13}\textbf{\color{blue}*} & \textbf{\color{blue}11.29} / \,\,\,\textbf{5.22} & \,\,\,\textbf{\color{blue}11.08} / \,\,\,\textbf{4.78}\textbf{\color{blue}*} & 
\textbf{18.61}\text{\color{blue}*} / \textbf{13.32}\,\,\, &
\,\,\,\text{\color{blue}\textbf{14.71}} / \,\,\,\textbf{8,49}\,\,\, & \textbf{\color{blue}13.57} / \,\,\,6.40
\\  \hline \hline

OFT-Net \cite{oft-net}  & Mono & \multirow{3}{*}{\makecell{Ped. \\ (IoU$\geq$0.5)}} &
\,\,\,\,\,\,-\,\,\,\,\,\, / \,\,\,1.11 &
\,\,\,\,\,\,-\,\,\,\,\,\, / \,\,\,1.06 &
\,\,\,\,\,\,-\,\,\,\,\,\, / \,\,\,1.06 &
\,\,\,\,\,\,-\,\,\,\,\,\, / \,\,\,1.55 &
\,\,\,\,\,\,-\,\,\,\,\,\, / \,\,\,1.93 &
\,\,\,\,\,\,-\,\,\,\,\,\, / \,\,\,1.65
\\

Linear System (Ours)  & Mono & &
\,\,\,1.51 / \,\,\,0.53 &
\,\,\,1.51 / \,\,\,0.53 &
\,\,\,1.51 / \,\,\,0.53 &
\,\,\,1.51 / \,\,\,0.53 &
\,\,\,1.51 / \,\,\,0.53 &
\,\,\,1.51 / \,\,\,0.53
\\

ShiftNet (Ours)  & Mono & &
\,\,\,\textbf{7.55} / \textbf{13.36} &
\,\,\,\textbf{6.80} / \textbf{10.59} &
\,\,\,\textbf{6.12} / \textbf{10.59} &
\,\,\,\textbf{8.24} / \textbf{13.81} &
\,\,\,\textbf{7.50} / \textbf{11.44} &
\,\,\,\textbf{6.73} / \textbf{10.76}
\\ \hline 

OFT-Net \cite{oft-net}  & Mono & \multirow{3}{*}{\makecell{Cyc. \\ (IoU$\geq$0.5)}} &
\,\,\,\,\,\,-\,\,\,\,\,\, / \,\,\,0.43 &
\,\,\,\,\,\,-\,\,\,\,\,\, / \,\,\,0.43 &
\,\,\,\,\,\,-\,\,\,\,\,\, / \,\,\,0.43 &
\,\,\,\,\,\,-\,\,\,\,\,\, / \,\,\,0.43 &
\,\,\,\,\,\,-\,\,\,\,\,\, / \,\,\,0.79 &
\,\,\,\,\,\,-\,\,\,\,\,\, / \,\,\,0.43 
\\

Linear System (Ours)  & Mono & &
\,\,\,1.38 / \,\,\,0.73 &
\,\,\,0.90 / \,\,\,0.43 &
\,\,\,0.90 / \,\,\,0.43 &
\,\,\,1.42 / \,\,\,0.53 &
\,\,\,0.90 / \,\,\,0.53 &
\,\,\,0.90 / \,\,\,0.53
\\

ShiftNet (Ours)  & Mono & &
\,\,\,\textbf{1.85} / \,\,\,\textbf{3.03} &
\,\,\,\textbf{1.08} / \,\,\,\textbf{3.03} &
\,\,\,\textbf{1.10} / \,\,\,\textbf{3.03} &
\,\,\,\textbf{2.30} / \,\,\,\textbf{3.58} &
\,\,\,\textbf{2.00} / \,\,\,\textbf{3.03} &
\,\,\,\textbf{2.11} / \,\,\,\textbf{3.03}
\\ \hline 
\end{tabular}}

\caption{3D object detection results on KITTI Chen/Test splits. We report $AP_{3D}(\%)$ and $AP_{BEV}(\%)$ for Car, Pedestrian and Cyclist classes. Methods that use a pre-trained monocular depth network (Mono+PD) are in {\color{blue}blue}. We denote with \textbf{\color{black}bold black} monocular state of the art, \textbf{\color{black}{bold black}\color{blue}*} cases when we outperform one Mono+PD method and \textbf{\color{blue}{bold blue}}, cases where we outperform all Mono+PD. Best viewed in color.}
\label{table:kitti_stats}
\end{table*}
\noindent\textbf{Volume Displacement Loss (VDL).}
Using an independent component-wise regression of $\mathbf{t}''$ does not capture correctly the final IoU detection measure. It does not take into account the object shape variations along the different dimensions. For example, an error along the Y-axis could produce a larger error in total volume than
one, by the same amount, along the Z-axis. A better loss is proposed by \cite{roi_10d}, by summing the Euclidean distances between each pair of the corresponding 3D bounding box corners, but without taking in consideration the different effects errors along the different axes might have, in the final IoU.

Based on this observation, we introduce the Volume Displacement Loss (VDL) to regress $\mathbf{t}''$ as the translation that optimizes the IoU score between two 3D boxes reconstructed using the same $\mathbf{d}$ and $\alpha_G$ parameters predicted in Stage 1, but one with the the ground-truth $\mathbf{t}$ and the other one, using the estimated $\mathbf{t}''$. While the direct, true 3D IoU w.r.t the ground truth bounding box is not differentiable and does not define a distance in metric space, we show that our VDL loss factorizes into a sum of simple terms and can be effectively used in training.

We first express the Signed Translation Displacement Error (STDE) vector $\Delta \mathbf{t} = [t''_x-t_x, t''_y-t_y, t''_z-t_z]^\top$ along the camera coordinates axis. We factor in the rotation $\alpha_G$ and decompose the STDE relative to the object's local coordinate axis, as it can be seen in the following Equation \ref{eq:rot_coord}:
\begin{equation}
\label{eq:rot_coord}
    \Delta \mathbf{t}_{\alpha_G} = \mathbf{R}_y(\alpha_G) \Delta \mathbf{t}  .
\end{equation}
The the VDL loss measures the displaced quantity of volume, when moving the object's center alongside one local coordinate axis. This quantity can be expressed as a function of the 3D bounding box dimensions and elements of  $\Delta \mathbf{t}_{\alpha_G} = [\Delta x_{\alpha_G}, \Delta y_{\alpha_G}, \Delta z_{\alpha_G}]^\top$. The following Equation \ref{eq:vdl_loss} defines our Volume Displacement Loss: 
\begin{equation}
\label{eq:vdl_loss}
     L = \underbrace{w \times h}_{A_{yOz}} \times |\Delta x_{\alpha_G}| +
                 \underbrace{w \times l}_{A_{xOz}} \times |\Delta y_{\alpha_G}| + 
                 \underbrace{h \times l}_{A_{xOy}} \times |\Delta z_{\alpha_G}|  .
\end{equation}

\section{Experiments}

In this section, we present the experimental setup, compare to 
state of the art monocular approaches, and test experimentally our contributions.

\subsection{Experimental Setup}
\noindent\textbf{Dataset.} We evaluate our architecture using the KITTI 3D Object Detection dataset \cite{kitti-3d}. We use the train-val split proposed by Chen \textit{et al.} \cite{mono3d}, which splits the KITTI training set into 3712 training images and 3769 validation images and also evaluate on the online test split. 
\noindent\textbf{Stage 1 Setup.} We use a Faster R-CNN architecture with ResNet-101 backbone, and the additional angle and 3D dimension heads as described in Section \ref{sec:stage1}. For ROI pooling we use image features from stride $8$ by atrous convolutions instead of strided ones in the final encoder part. Stage 1 network is trained 
end-to-end. At inference we use NMS for 2D boxes, angle, and dimension predictions.

\noindent\textbf{Stage 1 Training Procedure.} We train the model with SGD for 30 epochs with a batch size of 1, momentum of 0.9, initial learning rate of $3 \times 10^{ -4}$ and 2 learning rate decays of 0.1 at 1/2 and 2/3 of the epoch count. Faster R-CNN weights are pre-trained on the COCO 2D detection \cite{deataset:coco} and for the extra heads we use \textit{He initialization} \cite{he-initialization}. Our loss weighting strategy takes into account the uncertainty of a task: 1.0, 2.0, 5.0 and 100.0 for the classification, localization, orientation and dimension regression, respectively. We train the Stage 1 model on the KITTI dataset, using Chen's split \cite{mono3d} and the entire KITTI training data for the official submission. We train on the Car, Pedestrian, and Cyclist classes. We use random flip data augmentation.

\noindent\textbf{Stage 3 Setup.} 
We pre-train ShiftNet by using the corresponding ground-truth, instead of the Stage 1 generated data. This gives the network a good initialization on how to map ground truth 2D to 3D information. Then we fine-tune the network on real estimates produced by Stage 1. Compared to the Stage 2 least squares solution given by the mathematical system, ShiftNet improves on the 3D detection accuracy (0.7 3D IoU) on Car by 3\% on ground truth data (71\% vs. 68\%) and by 6.6\% on real images estimates given by the Faster R-CNN Stage 1 (13.8\% vs. 7.2\%). Our experiments show that a relatively simple neural net, such as ShiftNet, can infer 3D information even without directly using visual features, as can be seen in Fig. \ref{fig:ablation}. We experiment with augmenting the ground-truth with additional synthetic data, obtained by randomly perturbing the ground truth 3D boxes and then projecting them with the given camera matrix, to obtain 2D boxes information. The additional pre-training on synthetic data significantly improves the accuracy (0.5 3D IoU) for the classes with fewer examples (22\% vs. 20\% for Pedestrian, 73\% vs. 47\% for Cyclist), when training on ground-truth input.

Stage 1 and 3 models (Fig. \ref{fig:our_3stage_system}) are trained on a NVIDIA GTX 1080 Ti GPU and for one image, the total inference time is 259ms.

\subsection{Comparison to other Methods}
Our framework is evaluated on the KITTI $AP_{BEV}$ and $AP_{3D}$ evaluation metrics. We use all 3 classes: Car, Pedestrian, Cyclist. Each of the classes is split into 3 difficulties: \textit{Easy}, \textit{Medium}, and \textit{Hard}. In Table \ref{table:kitti_stats}, we compare with monocular 3D object detection methods. We show in \textit{blue} methods that use heavier monocular depth estimation networks. As we can see in Table \ref{table:kitti_stats}, it is clear that our method outperforms \textit{all} monocular 3D object detection methods that do not use depth estimation in $AP_{3D}$ and $AP_{BEV}$ on both \textit{val}/\textit{test} splits for \textit{Car} class. We outperform OFT-NET \cite{oft-net}, which uses \textit{orthographic feature transform}. We also outperform some methods that use monocular pre-trained depth prediction such as MLF-MONO \cite{mlf-mono} and ROI-10D \cite{roi_10d} on \textit{val} split, and come close to MLF-MONO on \textit{test} split. On KITTI \textit{test} split, these methods improve their results due to the much more training data available for depth prediction. We are also overall state of the art on the \textit{Pedestrian} and \textit{Cyclist} classes for monocular 3D detection. Thus, our result contradicts the assumption that these classes are too difficult to predict from monocular data.
\section{Discussion and Conclusions}
Compared to 2D detection (91\%) and object orientation (90\%), our $AP_{3D}$ is lower, due to noise in depth prediction, which, for an error of 1 meter, can move the 3D bounding box outside the required 0.7 3D IoU overlap. Overall, our approach proves very competitive, with top results in its category on the KITTI 3D Benchmark.
Our framework builds upon the robust Faster R-CNN, augmented with 3D object dimensions and local orientation prediction. We use the output of Faster R-CNN alongside the camera projection matrix to compute an analytic solution to the 3D translation problem. Our geometric solution is then refined with ShiftNet, which learns to leverage the output of Faster R-CNN with the rigorous mathematical solution. By combining the optimal, but less robust analytic solution with the imperfect, but much more robust deep learning model, we provide a hybrid system that takes advantage of the best of both worlds. With a much lighter architecture and without a pre-trained depth network we obtain state of the art results not only on the Car class, but also on the Pedestrian and Cyclist classes, on which we outperform our competitor OFT-NET \cite{oft-net} on all difficulty levels.                  

\bibliographystyle{ieee}
\bibliography{egbib}

\end{document}